\documentclass[11pt,a4paper]{article}
\usepackage[hyperref]{acl2020}
\usepackage{times}
\usepackage{latexsym}
\usepackage{graphicx}

\usepackage{microtype}

\aclfinalcopy 

\makeatletter
\newcommand\footnoteref[1]{\protected@xdef\@thefnmark{\ref{#1}}\@footnotemark}
\makeatother
\makeatletter
\newcommand{\printfnsymbol}[1]{%
  \textsuperscript{\@fnsymbol{#1}}%
}
\makeatother

\title{Sarcasm Detection using Context Separators in Online Discourse}
\author{
    Kartikey Pant \thanks{\hspace{2 mm} Both authors contributed equally to the work.} \and Tanvi Dadu \footnotemark[1] \\\
    International Institute of Information Technology, Hyderabad, India \\
    Netaji Subhas Institute of Technology, New Delhi, India \\
    {\tt kartikey.pant@research.iiit.ac.in} \\
    {\tt tanvid.co.16@nsit.net.in} \\
}

\date{}

\begin{document}
\maketitle
\begin{abstract}
Sarcasm is an intricate form of speech, where meaning is conveyed implicitly. Being a convoluted form of expression, detecting sarcasm is an assiduous problem. The difficulty in recognition of sarcasm has many pitfalls, including misunderstandings in everyday communications, which leads us to an increasing focus on automated sarcasm detection. In the second edition of the Figurative Language Processing (FigLang 2020) workshop, the shared task of sarcasm detection released two datasets, containing responses along with their context sampled from Twitter and Reddit. 

In this work, we use $RoBERTa_{large}$ to detect sarcasm in both the datasets. We further assert the importance of context in improving the performance of contextual word embedding based models by using three different types of inputs - \textit{Response-only}, \textit{Context-Response}, and \textit{Context-Response (Separated)}. We show that our proposed architecture performs competitively for both the datasets. We also show that the addition of a separation token between context and target response results in an improvement of $5.13\%$ in the \textit{F1-score} in the Reddit dataset.

\end{abstract}

\section{Introduction}

Sarcasm is a sophisticated form of speech, in which the surface meaning differs from the implied sense. This form of expression implicitly conveys the message making it hard to detect sarcasm in a statement. Since speech in sarcasm is dependent in context, it is tough to resolve the speaker’s intentions unless given insights into the circumstances of the sarcastic response. These insights or contextual information may include the speaker of the response, the listener of the response, and how its content relates to the preceding discourse.

Recognizing sarcasm is critical for understanding the actual sentiment and meaning of the discourse. The difficulty in the recognition of sarcasm causes misunderstandings in everyday communication. This difficulty also poses problems to many natural language processing systems, including summarization systems and dialogue systems. Therefore, it is essential to develop automated sarcasm detectors to help understand the implicit meaning of a sarcastic response.

The sarcasm detection shared task held at the second edition of the Figurative Language Processing (FigLang 2020) workshop proposes two datasets from different social media discourse platforms for evaluation. The first dataset contains user conversations from Twitter, while the second dataset contains Reddit conversation threads. Both datasets contain contextual information in the form of posts of the previous dialogue turns. Their primary aim is to understand the importance of conversational contextual information in improving the detection of sarcasm in a given response.

In this work, we explore the use of contextualized word embeddings for detecting sarcasm in the responses sampled from Reddit as well as Twitter. We outline the effect of adding contextual information, from previous dialogue turns, to the response, for both the datasets. We further explore the importance of separation tokens, differentiating discourse context from the target response while detecting sarcasm. 

\begin{figure*}[!h]
        \centering
        \includegraphics[width=16cm]{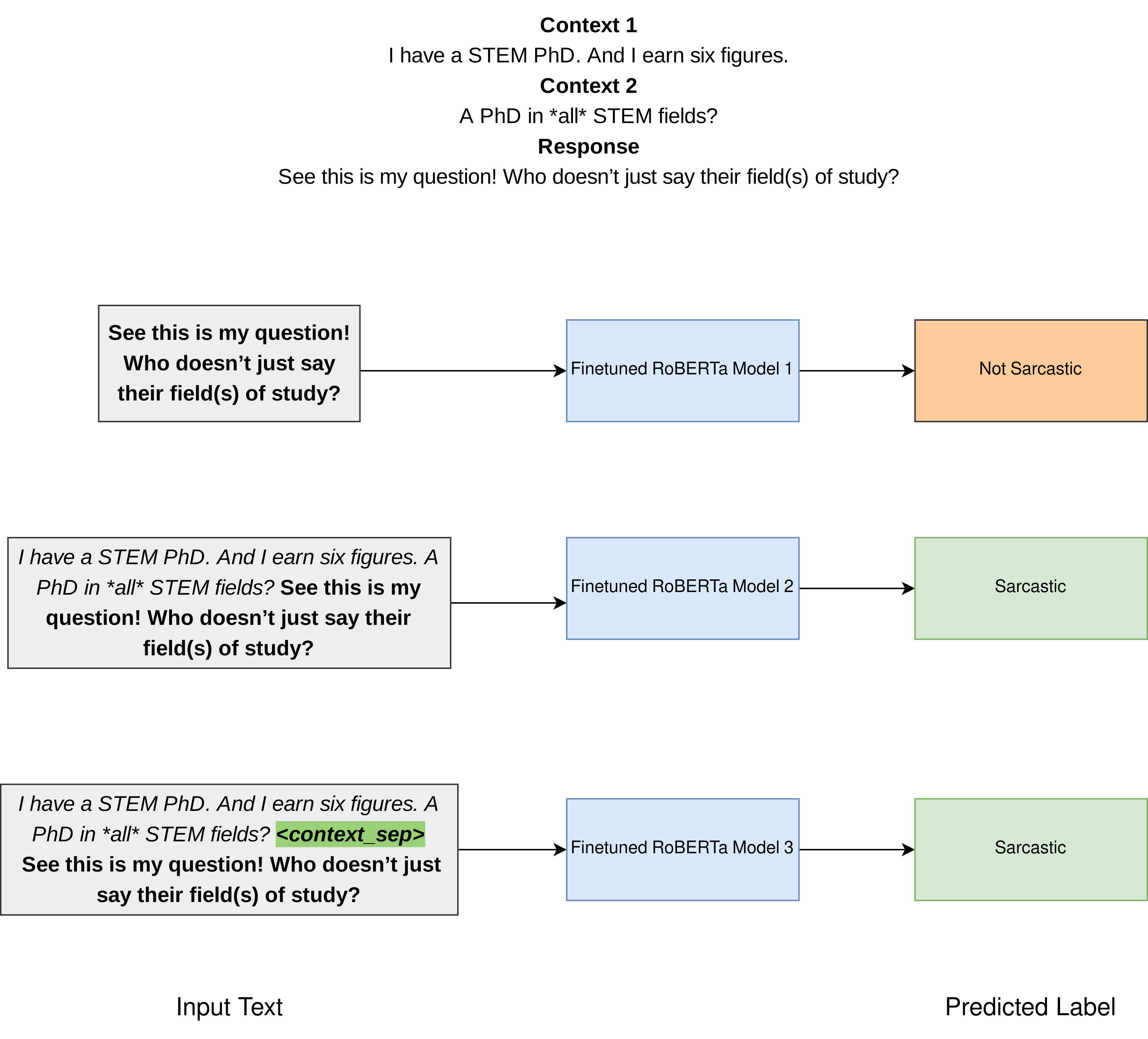}
        \caption{Architecture of our proposed approach.}
\end{figure*}

\section{Related Works}
\citet{davidov-etal-2010-semi} approached the task of sarcasm detection in a semi-supervised setting, investigating their algorithm in two different forms of text, tweets from Twitter, and product reviews from Amazon. Subsequently, \citet{gonzalez-ibanez-etal-2011-identifying} explored this task in a supervised setting, using SVM and logistic regression. They also release a dataset of 900 tweets to the public domain, entailing tweets containing sarcastic, positive, or negative content. Moreover, there has been significant work done to detect sarcasm in a multi-modal setting, primarily including visual cues. While \citet{Schifanella2016} proposed a multi-modal methodology to exploit features from both text and image, \citet{CaiSarcasm2019} investigated instilling attribute information from social media posts to propose a model leveraging hierarchical fusion.

\citet{Ghosh2018} investigated the use of Long Short-Term Memory (LSTM) networks with sentence-level attention for modeling both the responses under consideration and the conversation context preceding it. They execute their models on responses sampled from two social media platforms, Twitter and Reddit. They concluded that contextual information is vital for detecting sarcasm by showing that the LSTM network modeling the context and the response outperforms the LSTM network that models only the target response. This observation is in sync with other similar tasks like irony detection, as highlighted by \citet{Wallace2014HumansRC}, which claimed that human annotators consistently rely on contextual information to make judgments. 

The use of other turns of dialogue as contextual information has been explored well in previous works. \citet{Bamman-Smith-2015} investigated the use of “author and addressee” features apart from the conversation context, but observed only a minimal impact of modeling conversation context. \citet{oraby-et-al-2017} investigated using “pre” and “post” messages from debate forums and Twitter conversations to identify whether rhetorical questions are used sarcastically or not. They observe that using the “post” message as context improved the \textit{F1-score} for the sarcastic class. \citet{joshi-etal-2016-harnessing} showed that sequence labeling algorithms outperform traditional classical statistical methods, obtaining a gain of $4\%$ in \textit{F1-score} on their proposed dataset.

The use of pre-trained contextual word representations has been explored in multiple tasks in NLP, including text classification. Recently released models such as BERT \cite{Devlin2019} and RoBERTa \cite{2020roberta}, exploit the use of pre-training and bidirectional transformers to enable efficient solutions obtaining state-of-the-art performance. Pre-trained embeddings significantly outperform the previous state-of-the-art in similar problems such as humor detection \citep{Weller2019HumorDA}, and subjectivity detection \citep{subjective-detection}.

\section{Approach}
This section outlines the approach used to detect sarcasm in the Reddit and Twitter datasets. Our approach utilizes contextualized word embeddings with three different inputs, as explained in the following paragraphs. 

The use of pretrained contextualized word embeddings has been applied to achieve state-of-the-art results for various downstream tasks \cite{Devlin2019,2020roberta}. \citet{Devlin2019} proposed BERT that leverages context from both left and right representations in each layer of the bidirectional transformer. The model is pretrained and released in the public domain, and can be trained in a simpler, yet efficient manner without having to make significant architectural changes for a specific task.  

RoBERTa \citep{2020roberta} is a replication study of BERT, trained on a dataset twelve times larger with bigger batches as compared to BERT. RoBERTa makes use of larger byte-pair encoding(BPE) vocabulary that helps to achieve better results than BERT on various downstream tasks. We use $RoBERTa_{large}$ and finetune it for the task using varying hyperparameters for different inputs.

We use the following three different types of input to study the effect of context on the performance of $Roberta_{large}$ to detect sarcasm:
\begin{enumerate}
\item \textbf{Response-only} : Input containing only the target response.
\item \textbf{Context-Response} : Input containing target response appended to the related context (containing previous responses).
\item \textbf{Context-Response (Separated)}: Input containing a separation token separating target response from context (containing previous responses).
\end{enumerate}

\begin{table}[!htbp]
\centering
\begin{tabular}{|l|r|r|}
\hline
\textbf{Split/Dataset} & \textbf{Reddit} & \textbf{Twitter} \\ \hline
\textbf{Training} & 2.491 & 3.867 \\ \hline
\textbf{Testing} & 4.254 & 3.164 \\ \hline
\end{tabular}
\caption{The average number of contexts per post.}
\label{tab:avg-context-len}
\end{table}

\begin{table*}[!htbp]
\centering
\begin{tabular}{|l|r|r|r|}
\hline
\textbf{Input} & \textbf{F1-score} & \textbf{Precision} & \textbf{Recall} \\ \hline
\textbf{Response-only} & 0.752 & 0.752 & 0.753 \\ \hline
\textbf{Context-Response} & 0.772 & 0.772 & 0.772 \\ \hline
\textbf{Context-Response (Separated)} & 0.771 & 0.771 & 0.771 \\ \hline
\end{tabular}
\caption{Experimental Results for the Twitter test dataset.
}
\label{tab:twitter-results}
\end{table*}

\begin{table*}[!htbp]
\centering
\begin{tabular}{|l|r|r|r|}
\hline
\textbf{Input} & \textbf{F1-score} & \textbf{Precision} & \textbf{Recall} \\ \hline
\textbf{Response-only} & 0.679 & 0.679 & 0.679 \\ \hline
\textbf{Context-Response} & 0.681 & 0.684 & 0.692 \\ \hline
\textbf{Context-Response (Separated)} & 0.716 & 0.716 & 0.718 \\ \hline
\end{tabular}
\caption{Experimental Results for the Reddit test dataset.}
\label{tab:reddit-results}
\end{table*}

\section{Experiments}

\subsection{Dataset}

Two datasets containing an equal number of sarcastic and non-sarcastic responses were sampled from Twitter and Reddit by the authors of the shared task. The Twitter dataset comprises $5,000$ English tweets, and the Reddit dataset contains $4,400$ Reddit posts along with their context, which is an ordered list of dialogues. The sarcastic response, present in both datasets, is the reply to the last dialogue turn in the context list.

From Table \ref{tab:avg-context-len}, we infer that the dataset suffers from a significant mismatch in the average number of context responses between the training and testing split. This mismatch is particularly evident in the Reddit dataset, where the training split contains $1.71$ times the number of contexts provided on average as compared to the testing split. We observe a similar mismatch, in the opposite direction, with the testing split containing $1.22$ times the average number of contexts as compared to the training split. We argue that this mismatch in training and test splits in terms of context lengths, adds a layer of complexity to the problem. Consequently, we use the last two dialogues as the context in the \textit{Context-Response} input and \textit{the Context-Response (Separated)} input.

\subsection{Experimental Setting}
In this subsection, we outline the experimental setup for the sarcasm detection task and present the results obtained on the blind test set. For experiments, we used $RoBERTa_{large}$ having $355M$ parameters with a $50,265$ vocabulary size. For validation, we trained and evaluated our model for three different inputs using a $90-10$ train-validation split.

We finetune $RoBERTa_{large}$ with a learning rate of $1*10^{-5}$ for 3 epochs. We used a sequence length of $50$ for the \textit{Response-only} input and $256$ for the other two types of inputs, \textit{Context-Response} input and \textit{Context-Response (Separated)} input. We evaluate all our models on the following metrics: \textit{F1 score}, \textit{Precision-1} and \textit{Recall-1}.

\subsection{Results}

In Table \ref{tab:twitter-results} and Table \ref{tab:reddit-results}, we illustrate the effect of adding contextual information to the target response. We see an increase of $2.6\%$ and $0.3\%$ in \textit{F1-score} upon including previous contexts with the response in the Twitter dataset and Reddit dataset respectively. 

We also investigate the effect of adding a separation token between contextual information and the target response in the predictive performance of the model. We observe a $5.13\%$ increase in \textit{F1-score} in the Reddit dataset but a $0.1\%$ decrease in \textit{F1-score} in the Twitter dataset. We observe a similar pattern as the \textit{F1-score} in both of its constituent metrics of \textit{Precision} and \textit{Recall}.

\section{Conclusion}
This work presents the importance of context while detecting sarcasm from responses sampled from Twitter and Reddit. Our proposed architecture using three different inputs performs competitively for both datasets showing that the addition of contextual information to target response improves the performance of the fine-tuned contextual word embeddings for detecting sarcasm. We further show that the addition of a separation token between context and target response also performs competitively, markedly showing an improvement of $5.13\%$ in the \textit{F1-score} of Reddit dataset. Future works include exploring different contextual cues, including user-specific attribute information and extending this hypothesis to other figurative speeches like irony detection and humor detection.

\bibliography{anthology,acl2020}
\bibliographystyle{acl_natbib}

\appendix

\end{document}